\documentclass[11pt]{article}

\usepackage[preprint]{acl}

\usepackage{times}
\usepackage{latexsym}
\usepackage{url}
\usepackage[T1]{fontenc}

\usepackage[utf8]{inputenc}

\usepackage{microtype}

\usepackage{inconsolata}

\usepackage{graphicx}

\usepackage{url}
\usepackage{enumitem}
\usepackage{tabularx}
\usepackage[utf8]{inputenc} 
\usepackage[T1]{fontenc}    
\usepackage{url}            
\usepackage{booktabs}       
\usepackage{amsfonts}       
\usepackage{amsmath}
\usepackage{nicefrac}       
\usepackage{microtype}      
\usepackage{xcolor}         
\usepackage{colortbl}
\usepackage{graphicx}
\usepackage{svg}
\usepackage{multirow}
\usepackage{wrapfig}
\usepackage{threeparttable}
\usepackage{tikz}
\usepackage{makecell}
\usepackage{tablefootnote}

\usepackage{color}
\usepackage{xspace}
\usepackage{caption}
\usepackage{adjustbox}

\usepackage{subcaption}

\usepackage{amsmath}
\usepackage{mathtools}
\usepackage{amsthm}
\theoremstyle{plain}

\theoremstyle{definition}

\theoremstyle{remark}

\makeatletter
\def\BibTeXWarning#1{}
\makeatother

\newcommand{\mytitle}{How LLMs Fail and Generalize in RTL Coding for Hardware Design?}

\newcommand{\huckyang}[1]{{\color{blue}{\small\bf\sf [HY: #1]}}}
\newcommand{\guanting}[1]{{\color{orange}{\small\bf\sf [GT: #1]}}}
\newcommand{\Skip}[1]{}

\title{\mytitle}

\author{
  \textbf{Guan-Ting Liu\textsuperscript{1}},
  \textbf{Chao-Han Huck Yang\textsuperscript{1}},
  \textbf{Chenhui Deng\textsuperscript{1}},
  \textbf{Zhongzhi Yu\textsuperscript{1}},
\\
  \textbf{Brucek Khailany\textsuperscript{1}},
  \textbf{Yu-Chiang Frank Wang\textsuperscript{1}},
\\
  \textsuperscript{1}NVIDIA Research,
  \small{
    \textbf{Correspondence:} \href{mailto:dannliu@nvidia.com}{dannliu@nvidia.com}
  }
}

\begin{document}
\maketitle

\begin{abstract}
Translating sequential programming priors into the parallel temporal logic of hardware design remains a crucial bottleneck for large language models (LLM). To investigate this, we introduce a new error taxonomy grounded in problem solvability, inspired by cognitive theory. Our taxonomy categorizes failures into syntactic, semantic, solvable functional, and unsolvable functional types. Evaluations reveal a strict empirical ceiling on the VerilogEval benchmark, as frontier models plateau at a 90.8\% initial pass rate. These plateaus are defined by unsolvable functional errors, exposing persistent knowledge gaps immune to test time compute scaling. Furthermore, we expose a striking surface convergence gap: optimization readily eliminates syntax errors but concurrently exacerbates deeper functional failures. Our findings demonstrate that alignment techniques merely teach models to compile. While repeated sampling strategies can patch solvable errors, register-transfer level (RTL) coding capacity\footnote{Our demo: \url{huggingface.co/spaces/nvidia/LLM_RTL_Errors_Explainer}} remains strictly bounded by pretraining knowledge. Addressing challenges in the current LLM based hardware generation pipeline requires more studies in model reasoning rather than alignment interventions. 

\end{abstract}
\section{Introduction}
\label{sec:intro}

\begin{figure}[t]
\centering
\includegraphics[trim={0 4.2cm 6.5cm 0},width=\columnwidth]{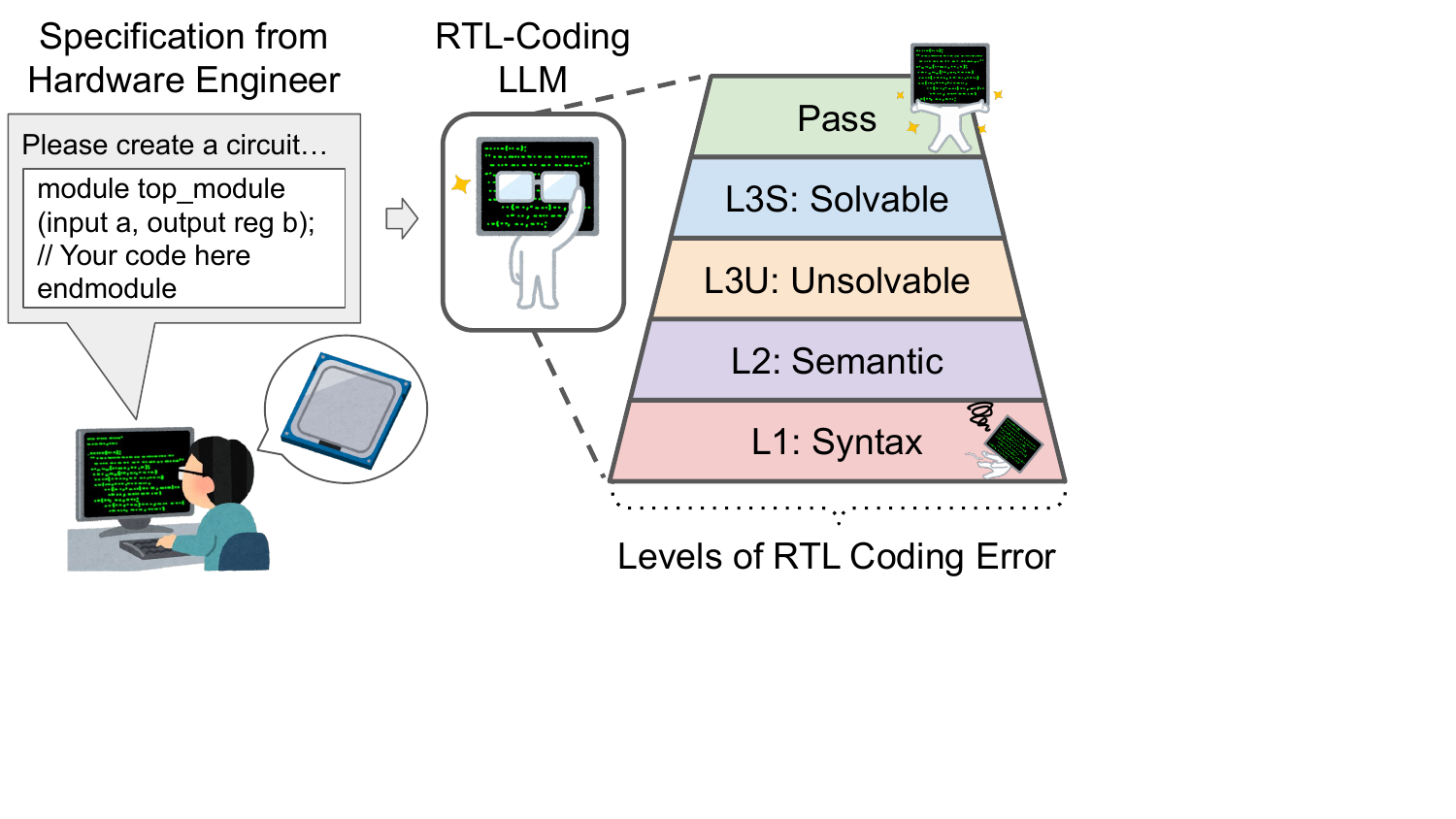}
\caption{Four-level error taxonomy for LLM-generated RTL code. Each rollout is classified by successive stages of hardware design: parsing (L1), compilation/synthesis (L2), and testbench simulation (L3). L3 failures are further split by \emph{problem-level solvability}: L3S if any rollout for the same problem passes the testbench (addressable via inference-time scaling), L3U if none passes (requires model improvement). Examples are shown in Table~\ref{tab:error_examples}.}
\label{fig:taxonomy_overview}
\end{figure}

Large language models (LLMs) have achieved substantial gains on multiple domains, including mathematics~\cite{shao2024deepseekmath}, general reasoning~\cite{deepseek2025r1,wei2022chain}, and code generation~\cite{le2022coderl,chen2021evaluating}.
However, for challenging domains like hardware design that require reasoning, and domain-specific knowledge, languages models often fail or achieve limited success through cognitive distortion~\citep{wan2025speechiq}.

In the domain of hardware design, engineers use hardware description languages (HDLs)~\cite{thomas2002verilog} such as Verilog to define circuits at the Register-Transfer Level (RTL), an abstraction that models the flow of digital signals between hardware registers and the logical operations performed on those signals.
One key challenge for LLMs learning RTL coding is that, unlike C or Python where statements execute sequentially, RTL code describes \emph{parallel} signal flows: all logical operations on each signal path may happen independently and simultaneously.
This parallelism is difficult to trace from a sequential-execution perspective and prone to race conditions~\cite{cummings2000nonblocking}.
Another challenge is that RTL code is far less represented on the internet compared to C or Python, making it difficult to collect high-quality, comprehensive training data.

In face of these challenges, the two key question we want to investigate are: \textbf{can LLMs leverage what they have learned during pretraining to solve RTL coding tasks, and can Reinforcement Learning (RL) fine-tuning help them generalize prior knowledge from other domains to the hardware domain?}
To solve these questions, a growing body of work~\cite{liu2024craftrtl,liu2024rtlcoder,zhao2024codev,deng2025scalertl,deng2026acertl,zhu2026qimengcodevr1} has finetuned LLMs for better RTL-coding capabilities.
Together with prior RTL-specialized approaches, we evaluate proprietary frontier models and open-source models on VerilogEval-Human benchmark~\cite{liu2023verilogeval} to understand how different model families perform on these challenging tasks.
Table~\ref{tab:error_breakdown} summarizes the error breakdown across all evaluated models, and we find that common error types and fault patterns recur across results from very different models.

To understand \emph{when and why} language models fail on RTL coding, we perform a detailed analysis and propose a four-level error taxonomy: L1~syntactic, L2~semantic, and L3~functional discrepancies further split by \emph{problem-level solvability} into L3S~(Solvable) and L3U~(Unsolvable). The illustration and formal definition of the taxonomy can be find in Figure~\ref{fig:taxonomy_overview} and Section~\ref{sec:fault}.
With the proposed error type taxonomy, we can identify how LLMs fail RTL-coding tasks from perspectives of syntax, semantic and functional correctness as shown in Table~\ref{tab:error_examples}.
More importantly, we can measure the effectiveness of supervised fine-tuning (SFT) or RL fine-tuning with the proposed taxonomy to understand the before/after failing behavior (Figure~\ref{fig:sft_transition_matrix}, Figure~\ref{fig:transition_matrix}) and improve the training strategy accordingly. Further discussion about the learning dynamic and failing behavior analysis across proprietary and open-source models can be found in Section~\ref{sec:prior_fault} and Section~\ref{sec:experiments}.

\begin{table}[t!]
\centering
\caption{Error breakdown (\%) on VerilogEval-Human v1 benchmark (156 problems, 10 rollouts each).
L1+L2+L3S+L3U+Pass=100\%.
All evaluation is conducted in a single-turn chat session without tool calling, compilation feedback or iterative refinement.}
\label{tab:error_breakdown}
\footnotesize
\setlength{\tabcolsep}{4pt}
\begin{tabular}{l ccccc}
\toprule
Model & L1 & L2 & L3S & L3U & Pass@1 \\
\midrule
\multicolumn{6}{l}{\textit{Proprietary}} \\
~~Claude Opus 4.6 & 2.4 & 0.3 & 2.2 & 4.2 & 90.8 \\
~~GPT-5.3 Codex & 2.7 & 0.6 & 2.7 & 5.0 & 89.0 \\
~~Gemini 3.1 Pro & 8.4 & 0.5 & 1.2 & 3.6 & 86.3 \\
~~GPT-5.4 & 1.2 & 0.6 & 6.6 & 10.0 & 81.7 \\
~~GPT-5.2 & 0.0 & 7.6 & 4.6 & 11.0 & 76.9 \\
~~Claude Sonnet 4.6 & 11.3 & 0.3 & 5.6 & 6.7 & 76.2 \\
~~GPT-OSS-120B & 12.2 & 3.3 & 8.8 & 6.6 & 69.0 \\
~~GPT-5.1 & 6.1 & 4.0 & 7.0 & 15.0 & 67.9 \\
~~Gemini 3 Pro & 29.3 & 0.1 & 2.0 & 4.2 & 64.4 \\
\midrule
\multicolumn{6}{l}{\textit{RTL-special.}} \\
~~CodeV-R1-Distill-7B & 2.5 & 2.7 & 11.8 & 16.7 & 66.3 \\
~~CodeV-R1-Qwen-7B & 1.1 & 2.1 & 11.5 & 15.6 & 69.7 \\
~~ScaleRTL-32B & 1.5 & 1.5 & 12.0 & 10.0 & 75.0 \\
\midrule
\multicolumn{6}{l}{\textit{Open-source}} \\
~~Qwen2.5-Coder-7B & 57.4 & 6.5 & 4.4 & 19.7 & 11.9 \\
~~Qwen2.5-Coder-32B & 56.5 & 4.6 & 2.6 & 20.9 & 15.4 \\
~~DS-R1-Distill-32B & 24.7 & 11.0 & 5.3 & 10.0 & 49.0 \\
~~K2-Think-SFT & 15.4 & 6.9 & 4.0 & 9.1 & 64.5 \\
~~K2-Think & 12.3 & 6.5 & 5.7 & 8.3 & 67.1 \\
\midrule
\multicolumn{6}{l}{\textit{RL-finetuned}} \\
~~K2-Think-SFT & 7.4 & 4.2 & 6.4 & 10.2 & 71.8 \\
~~K2-Think & 7.8 & 2.5 & 7.1 & 9.6 & 73.1 \\
\bottomrule
\end{tabular}
\end{table}

To further observe how models can \emph{generalize} their prior knowledge to RTL through reinforcement learning, we conduct GRPO~\cite{shao2024deepseekmath} training on two open-source K2-Think models and perform fault analysis on the generated rollouts throughout training (Section~\ref{sec:experiments}).
Since the K2 models we study are pretrained and instruction-tuned using exclusively non-RTL data, our GRPO fine-tuning on RTL tasks constitutes a \emph{cross-domain generalization} process---testing whether RL can transfer coding competencies acquired from general-purpose programming to RTL coding in hardware design tasks.

Based on these analyses, we conclude:

\begin{enumerate}[leftmargin=*,itemsep=1pt]
    \item SFT and RL reduce L1+L2 errors but \emph{increase} L3 failures---models learn to compile without holistic hardware understanding, revealing pre-existing knowledge gaps at the verification (testbench) stage (Figure~\ref{fig:sft_transition_matrix}, Figure~\ref{fig:transition_matrix}). 
    \item L3U~(Unsolvable) errors persist as an irreducible ceiling (4--17\%) across all training paradigms; even the best frontier model plateaus at 90.8\% pass with a 4.2\% L3U floor (Table~\ref{tab:error_breakdown} and Section~\ref{sec:prior_fault}).
    \item The L3S/L3U split is actionable: L3S errors are recoverable via best-of-$N$ sampling, while L3U errors require model improvement. RTL-specialized models have the highest L3S rates (11--12\%), benefiting most from inference-time scaling (Section~\ref{sec:prior_fault}, Section~\ref{sec:solvability}).
    \item Most L3U errors are model-specific, not intrinsically unsolvable: the union of all 17 evaluated models solves 150/156 problems (96.2\%), yet no single model exceeds 148---the gap reveals complementary knowledge across model families, suggesting that model diversity can nearly eliminate functional failures (Section~\ref{sec:prior_fault}).

\end{enumerate}

\section{Related Work}
\label{sec:related}

\begin{table*}[t]
\centering
\caption{Representative Verilog code examples for each error level in the four-level taxonomy. The \textbf{Problem Spec} row shows the task description and module header provided to the LLM; the remaining rows show the generated body and its correction. Errors are highlighted in \textcolor{red}{red}; fixes in \textcolor{green!50!black}{green}.}
\label{tab:error_examples}
\small
\setlength{\tabcolsep}{4pt}
\renewcommand{\arraystretch}{1.15}
\begin{tabularx}{\textwidth}{@{} p{1.5cm} X X X X @{}}
\toprule
& \textbf{L1 (Syntactic)}
& \textbf{L2 (Semantic)}
& \textbf{L3S (Solvable)}
& \textbf{L3U (Unsolvable)} \\
\midrule
\textbf{\newline Problem Spec}
& 2-to-1 multiplexer. \newline
  \texttt{module mux2(} \newline
  \texttt{~~input a,b,sel,} \newline
  \texttt{~~output reg out);}
& Positive-edge D flip-flop. \newline
  \texttt{module dff(} \newline
  \texttt{~~input clk, d,} \newline
  \texttt{~~output q);}
& BCD counter (counts \newline 0--9, then wraps to 0). \newline
  \texttt{module bcd(} \newline
  \texttt{~~input clk, rst,} \newline
  \texttt{~~output reg [3:0] cnt);}
& 4-bit LFSR with polynomial $x^4\!+\!x^3\!+\!1$. \newline
  \texttt{module lfsr4(} \newline
  \texttt{~~input clk, rst,} \newline
  \texttt{~~output reg [3:0] q);} 
\\
\midrule
\textbf{\newline \newline LLM \newline Output} 
& \texttt{always @(*) begin} \newline
  \texttt{out = sel ? b : a} \newline
  \texttt{\textcolor{red}{/* no end */}} \newline
  \texttt{endmodule}
& \texttt{\textcolor{red}{/* no reg decl. */}} \newline
  \texttt{always @(posedge clk)} \newline
  \texttt{~~\textcolor{red}{q} <= d;} \newline
  \texttt{endmodule}
& \texttt{always @(posedge clk)} \newline
  \texttt{~~if (rst) cnt<=0;} \newline
  \texttt{~~else if} \newline
  \texttt{~~~~(\textcolor{red}{cnt >= 10}) cnt<=0;} \newline
  \texttt{~~else cnt<=cnt+1;} \newline
  \texttt{endmodule}
& \texttt{always @(posedge clk)} \newline
  \texttt{~~if (rst) q<=4'b1;} \newline
  \texttt{~~else} \newline
  \texttt{~~~~\textcolor{red}{q<=\{q[2:0],}} \newline
  \texttt{~~~~\textcolor{red}{~~~~~q[3]\};}} \newline
  \texttt{endmodule}
\\
\midrule
\textbf{\newline \newline Corrected}
& \texttt{always @(*) begin} \newline
  \texttt{~~out = sel ? b : a\textcolor{green!50!black}{;}} \newline
  \texttt{\textcolor{green!50!black}{end}} \newline
  \texttt{endmodule}
& \texttt{\textcolor{green!50!black}{reg q;}} \newline
  \texttt{always @(posedge clk)} \newline
  \texttt{~~q <= d;} \newline
  \texttt{endmodule}
& \texttt{always @(posedge clk)} \newline
  \texttt{~~if (rst) cnt<=0;} \newline
  \texttt{~~else if} \newline
  \texttt{~~~~(\textcolor{green!50!black}{cnt == 9}) cnt<=0;} \newline
  \texttt{~~else cnt<=cnt+1;} \newline
  \texttt{endmodule}
& \texttt{always @(posedge clk)} \newline
  \texttt{~~if (rst) q<=4'b1;} \newline
  \texttt{~~else} \newline
  \texttt{~~~~\textcolor{green!50!black}{q<=\{q[2:0],}} \newline
  \texttt{~~~~\textcolor{green!50!black}{~q[3]\^{}q[2]\};}} \newline
  \texttt{endmodule}
\\
\bottomrule
\end{tabularx}
\end{table*}

\paragraph{RTL Code Generation with LLMs.}
LLM-based RTL code generation has grown rapidly, spanning benchmarks~\cite{liu2023verilogeval,lu2024rtllm}, domain-adapted models~\cite{liu2024rtlcoder,deng2025scalertl,liu2024chipnemo,liu2024craftrtl,zhao2024codev}, and RL-based training~\cite{zhu2026qimengcodevr1}.
VerilogEval~\cite{liu2023verilogeval} provides the standard evaluation framework for Verilog generation, while RTLLM~\cite{lu2024rtllm} extends this to more complex designs.
RTLCoder~\cite{liu2024rtlcoder}, CodeV~\cite{zhao2024codev}, ScaleRTL~\cite{deng2025scalertl}, and CraftRTL~\cite{liu2024craftrtl} construct high-quality training corpora for domain adaptation. 
ACE-RTL~\cite{deng2026acertl} augments RTL models with agentic iterative refinement.
CodeV-R1~\cite{zhu2026qimengcodevr1} applies adaptive DAPO~\cite{yu2025dapo} as RLVR (Reinforcement Learning with Verifiable Rewards) to RTL tasks, reporting significant pass-rate improvements.

\paragraph{RL for Code Generation.}
CodeRL~\cite{le2022coderl} pioneered RL fine-tuning for code with execution-based rewards.
Subsequent work has explored execution feedback~\cite{shojaee2023execution}, process-reward models~\cite{lightman2023lets}, and GRPO~\cite{shao2024deepseekmath,deepseek2025r1} for reasoning tasks.
DeepSeek-R1~\cite{deepseek2025r1} demonstrates that GRPO can elicit chain-of-thought reasoning without supervised data.
However, these studies focus on general-purpose code or mathematics; the dynamics of RL fine-tuning for hardware description languages remain under-explored.

\paragraph{Error Analysis of Code LLMs.}
Systematic error analysis of code LLMs is relatively sparse.
\citet{yasunaga2021break} study unsupervised program repair, and \citet{huang2024large} show LLMs cannot self-correct reasoning.
InferFix~\cite{jin2023inferfix} analyses LLM-assisted program repair.
~\citet{zhang2025understanding} manually categorise 306 faulty Verilog designs from five LLMs by root cause, providing a complementary perspective through manual inspection.
Our work differs in providing 4-level error taxonomy that is applicable to generic RTL coding task without manual inspection or modification. 

\section{Fault Analysis Framework}
\label{sec:fault}

We introduce a fault analysis framework comprising a four-level error taxonomy grounded in the HDL compilation and verification pipeline: L1~syntactic, L2~semantic, L3S~functional-solvable, and L3U~functional-unsolvable (Figure~\ref{fig:taxonomy_overview}).
This framework is applicable to any LLM generating Verilog/SystemVerilog code evaluated against testbenches.

\subsection{Preliminaries and Notation}
\label{sec:prelim}

Let $S$ denote a source code string produced by an LLM in response to an RTL design specification.
Let $G$ be the formal grammar of the target hardware description language (IEEE~1364 for Verilog, IEEE~1800 for SystemVerilog), and let $L(G)$ denote the language generated by $G$---the set of all syntactically valid strings.
When $S \in L(G)$, the parser constructs an abstract syntax tree $\mathrm{AST}(S)$.

Beyond syntactic validity, HDL toolchains enforce a set of \emph{static semantic constraints} $C = \{c_1, \ldots, c_m\}$, where each $c_i : \mathrm{AST} \to \{\textit{True}, \textit{False}\}$ is a predicate evaluated on the abstract syntax tree during elaboration, linting, or synthesis.
We say that $S$ \emph{satisfies} $C$ when $\forall c \in C,\; c(\mathrm{AST}(S)) = \textit{True}$, and that $S$ \emph{violates} $C$ when $\exists\, c \in C : c(\mathrm{AST}(S)) = \textit{False}$.

Finally, let $M_S$ denote the structural hardware model synthesised from $S$, let $\Phi$ denote the formal design specification (expressed as testbench assertions), and let $\models$ denote the satisfaction relation: $M_S \models \Phi$ iff $M_S$ satisfies every assertion in $\Phi$.

\paragraph{Verilog-Specific Constraint Instances.}
For Verilog/SystemVerilog, $C$ includes the following widely enforced constraints:
\begin{itemize}[leftmargin=*,itemsep=1pt]
    \item $c_{\text{type}}$ \textbf{(Procedural Type)}: nets (\texttt{wire}) cannot be targets in procedural blocks; registers (\texttt{reg}) cannot be targets of continuous assignments.
    \item $c_{\text{sync}}$ \textbf{(Synchronous Assignment)}: non-blocking assignments (\texttt{<=}) for sequential logic in clocked \texttt{always} blocks; blocking (\texttt{=}) for combinational logic~\cite{cummings2000nonblocking}.
    \item $c_{\text{sens}}$ \textbf{(Combinational Sensitivity)}: combinational \texttt{always} blocks must be sensitive to all signals read within the block.
    \item $c_{\text{latch}}$ \textbf{(Combinational Completeness)}: all branches of conditional statements in combinational blocks must assign every output, to avoid unintentional latch inference.
    \item $c_{\text{drive}}$ \textbf{(Multiple Driver)}: each net may have at most one continuous driver; each register may be assigned in at most one \texttt{always} block.
    \item $c_{\text{port}}$ \textbf{(Instantiation Port)}: module instantiations must match port declarations in name, direction, and width.
    \item $c_{\text{synth}}$ \textbf{(Synthesizability)}: code must avoid unsynthesisable constructs (e.g., \texttt{\#} delays, dynamic memory) for synthesis targets.
\end{itemize}

\subsection{Formal Error Taxonomy}
\label{sec:error_def}

Using the notation above, we define three mutually exclusive error levels that, together with a passing outcome, exhaustively partition every rollout.

\paragraph{Definition 1 (Syntactic Discrepancy, $E_{\text{syn}}$).}
\vspace{-4pt}
\begin{equation}
E_{\text{syn}} \iff S \notin L(G)
\end{equation}
The source string is rejected by the HDL parser.  No AST can be constructed; the design cannot proceed to elaboration.  Detected by the \emph{lexer/parser} stage.

\paragraph{Definition 2 (Semantic Discrepancy, $E_{\text{sem}}$).}
\vspace{-4pt}
\begin{equation}
\begin{split}
E_{\text{sem}} \iff{}& \bigl(S \!\in\! L(G)\bigr) \\
  &\land\, \bigl(\exists\, c \!\in\! C : c(\mathrm{AST}(S)) = \textit{False}\bigr)
\end{split}
\end{equation}
The source string parses into a valid AST but violates at least one static semantic constraint.  Detected during \emph{elaboration}, \emph{linting}, or \emph{synthesis}.

\paragraph{Definition 3 (Functional Discrepancy, $E_{\text{fun}}$).}
\vspace{-4pt}
\begin{equation}
\begin{split}
E_{\text{fun}} \iff{}& \bigl(S \!\in\! L(G)\bigr) \\
  &\land\, \bigl(\forall c \!\in\! C,\, c(\mathrm{AST}(S)) = \textit{True}\bigr) \\
  &\land\, \bigl(M_S \!\not\models\! \Phi\bigr)
\end{split}
\end{equation}
The source string parses and satisfies all static constraints, but the synthesised model fails to meet the design specification.  Detected during \emph{simulation/verification}.

\medskip\noindent
We further decompose functional discrepancies by \emph{problem-level solvability}For instance, whether the model demonstrates the ability to solve the problem in any rollout.

To define solvability we introduce rollout-set notation.
Let $\mathcal{R}_i = \{r_1, \ldots, r_K\}$ be the set of $K$ rollouts sampled for problem~$i$, and let $\mathrm{pass}(r)$ indicate whether rollout $r$ passes \emph{all} assertions during testbench simulation.

\paragraph{Definition 3a (Solvable Functional, $E_{\text{sol}}$).}
\vspace{-4pt}
\begin{equation}
\begin{split}
E_{\text{sol}}(r) \iff{}& E_{\text{fun}}(r) \\
  &\land\, \bigl(\exists\, r' \!\in\! \mathcal{R}_i : \mathrm{pass}(r')\bigr)
\end{split}
\end{equation}
The model \emph{can} solve the problem (proven by at least one passing rollout), but this particular rollout fails.
A stochastic failure addressable via inference-time compute scaling (best-of-$N$ sampling).

\paragraph{Definition 3b (Unsolvable Functional, $E_{\text{unsol}}$).}
\vspace{-4pt}
\begin{equation}
\begin{split}
E_{\text{unsol}}(r) \iff{}& E_{\text{fun}}(r) \\
  &\land\, \bigl(\forall\, r' \!\in\! \mathcal{R}_i : \lnot\mathrm{pass}(r')\bigr)
\end{split}
\end{equation}
No rollout passes; the model lacks the knowledge to solve this problem.
A true knowledge gap that requires model improvement through better pretraining data, fine-tuning, or architectural changes.

\paragraph{Mutual Exclusivity and Exhaustiveness.}
The two sub-levels partition $E_{\text{fun}}$ by a single binary axis---solvability---determined by whether any rollout in $\mathcal{R}_i$ passes:
\begin{equation}
E_{\text{sol}} \,\lor\, E_{\text{unsol}} \iff E_{\text{fun}}
\end{equation}
Combined with L1 and L2, the full taxonomy yields four mutually exclusive error outcomes (plus Pass) for every rollout.

\medskip\noindent
We abbreviate $E_{\text{syn}}$, $E_{\text{sem}}$, $E_{\text{sol}}$, $E_{\text{unsol}}$ as \textbf{L1}, \textbf{L2}, \textbf{L3S}, \textbf{L3U} respectively in tables and figures, and use \textbf{L3} as shorthand for the union $\text{L3S} + \text{L3U}$.
At every evaluation point, the following invariant holds:
\[
\textbf{L1 + L2 + L3S + L3U + Pass = 100\%}
\]
Syntactic discrepancies (L1) include keyword hallucinations, missing module boundaries, and malformed expressions.
Semantic discrepancies (L2) encompass violations such as wire targets in \texttt{always} blocks ($c_{\text{type}}$), blocking/non-blocking confusion ($c_{\text{sync}}$), incomplete sensitivity lists ($c_{\text{sens}}$), unintentional latch inference ($c_{\text{latch}}$), multiple drivers ($c_{\text{drive}}$), port mismatches ($c_{\text{port}}$), and unsynthesisable constructs ($c_{\text{synth}}$).
Solvable functional errors (L3S) represent problems where the model succeeds in at least one rollout but fails on this particular attempt---the model possesses the requisite knowledge but does not consistently apply it.
Unsolvable functional errors (L3U) represent problems where the model fails across all $K$ rollouts---a persistent knowledge gap that no amount of additional sampling can resolve.


\subsection{Fault Analysis Across Model Families}
\label{sec:prior_fault}

To understand how current language models perform on RTL coding tasks and where they systematically fail, we evaluate nine proprietary frontier models, five open-source models, and three RTL-specialized models on VerilogEval-Human with 10 rollouts per problem (temperature 0.2).
All evaluations are single-turn: each rollout is an independent generation without compilation feedback or iterative refinement.
Table~\ref{tab:error_breakdown} presents the error breakdown (L1~syntactic, L2~semantic, L3S~solvable functional, L3U~unsolvable functional) for all evaluated models.
The RL fine-tuning experiments in Section~\ref{sec:experiments} complement this analysis by studying whether models can generalize prior knowledge from other domains to RTL.
We organise the findings into within-type and before-after comparisons.

\paragraph{(1)~Within proprietary: distinct failure profiles.}
Frontier models span a 26.4~pp range in pass@1 accuracy (64.4--90.8\%) with strikingly different bottlenecks.
\emph{Syntax-bottlenecked} models such as Gemini~3~Pro (L1=29.3\%, but L3U only 4.1\%) lose most rollouts before functional testing even begins.
In contrast, \emph{functional-bottlenecked} models such as GPT-5.4 (L1=1.2\%, L3U=10.0\%) produce compilable code that fails at the testbench---their failures are deep knowledge gaps, not format issues.
GPT-5.2 is a unique outlier: L1=0.0\% (perfect syntax) but L2=7.6\%---the only frontier model with significant semantic errors---suggesting it masters HDL grammar but misuses constructs such as blocking vs.\ non-blocking assignments.


\paragraph{(3)~Within RTL-specialized: high L3S rates reveal inference-time scaling opportunity.}
All three RTL-specialized models have very low L1+L2 ($<$5\%) but the \emph{highest} L3S rates of any group (11.5--12.0\% vs.\ 1.2--8.8\% for frontier), meaning they solve many problems inconsistently across rollouts.
Since L3S errors are recoverable via best-of-$N$ sampling, RTL-specialized models paradoxically benefit \emph{more} from inference-time compute scaling than frontier models.
ScaleRTL-Qwen-32B (L3U=10.0\%) has substantially lower L3U than CodeV-R1-Distill-7B (L3U=16.7\%), confirming that the 32B base model contributes better latent functional knowledge that SFT can surface.

\begin{figure}[t]
\centering
\includegraphics[width=\columnwidth]{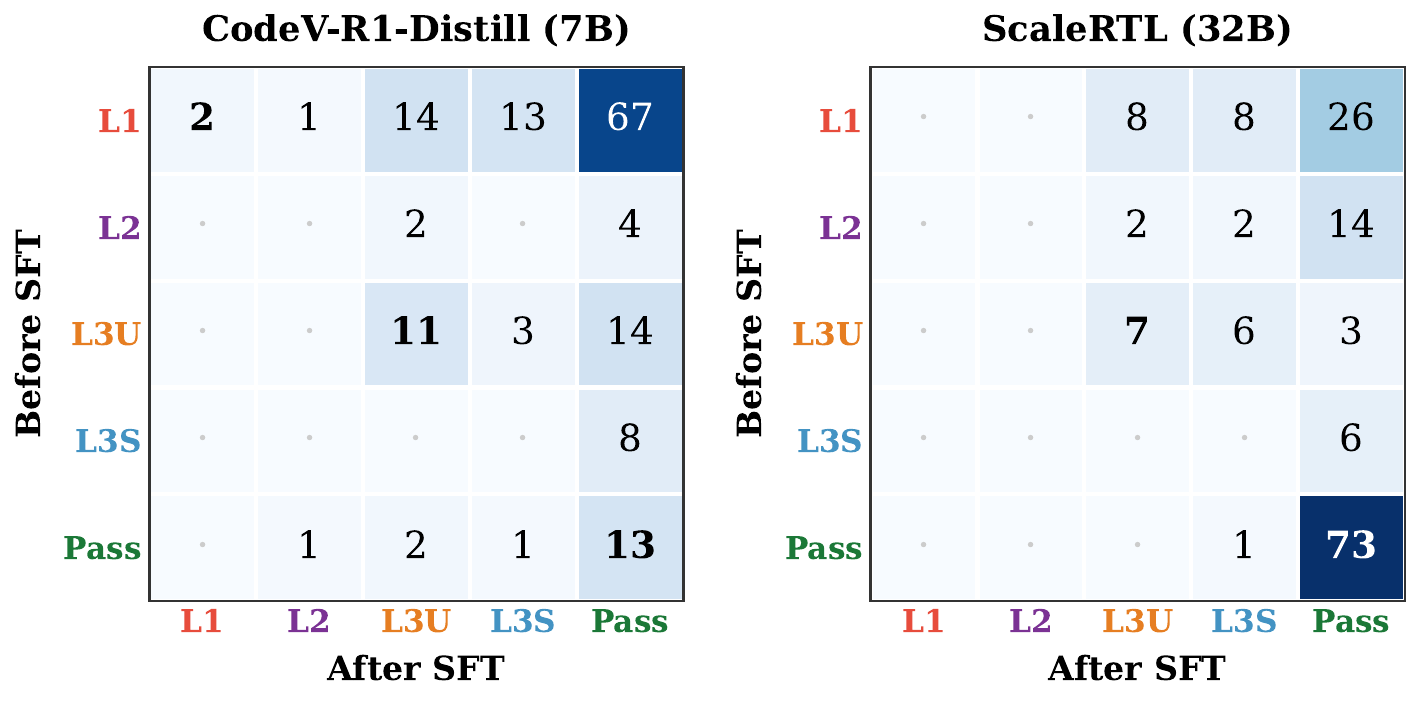}
\caption{Problem-level transition matrices (before vs.\ after SFT). Each of the 156 VerilogEval problems is assigned a majority-vote category over its 10 rollouts. Cell~$(i,j)$ counts problems moving from category~$i$ to~$j$. At 7B, SFT acts as a ``syntax rescue'' (67/97 L1$\to$Pass). At 32B, SFT completely eliminates L1 and L2 while preserving 73/74 existing Pass. Only 4 and 1 relapses (Pass$\to$L1/L2/L3U/L3S) at 7B and 32B respectively, with persistent L3U as the hard ceiling.}
\label{fig:sft_transition_matrix}
\end{figure}

\paragraph{(4)~Before-after: SFT moves errors forward.}
Comparing each base model to its RTL-specialized derivative reveals a consistent \emph{error pipeline progression}: SFT converts compilation failures into either Pass or L3 functional failures.
From the base model (Qwen2.5-Coder-7B) to RTL-specialized model (CodeV-R1-Distill-7B), SFT drops compilation errors (L1+L2) by 58.7~pp, raises Pass by 54.4~pp, but total L3 \emph{increases} by 4.4~pp.
At 32B scale, DS-R1-Distill-32B (L1+L2=35.7\%, Pass=49.0\%) becomes ScaleRTL-32B (L1+L2=3.0\%, Pass=75.0\%)---SFT drops compilation errors by 32.7~pp, raises Pass by 26.0~pp, but total L3 \emph{increases} by 6.7~pp (15.3\%$\to$22.0\%).

To track how individual problems move between error levels, we assign each of the 156 problems a majority-vote category over its 10 rollouts (with lower error level as tie breaker). The resulting problem-level transition matrices are showin in Figure~\ref{fig:sft_transition_matrix}).
At the 7B scale, SFT acts primarily as a \textit{syntax rescue}: 67 of the 97 L1 problems jump directly to Pass, while another 27 advance to L3 (13 L3S, 14 L3U), leaving only two problems at L1.
At 32B, SFT achieves \emph{complete elimination} of both L1 and L2 (42$\to$0 and 18$\to$0), while 73 of 74 already-passing problems remain stable.
Critically, L3U is the stickiest category: 11 of 28 L3U problems at 7B and 7 of 16 at 32B persist through SFT, confirming that L3U represents a hard knowledge ceiling that domain-specific fine-tuning alone cannot breach.
Both scales show exactly seven regressions, defined as previously passing problems that SFT destabilizes, suggesting a small but consistent cost of adaptation. SFT primarily teaches the model to produce syntactically and semantically valid Verilog; it does not instil new hardware understanding but merely allows existing knowledge (or lack thereof) to be revealed at the functional testing stage.

\paragraph{(6)~Before-after: RL increases solvability but exposes knowledge gaps.}
Comparing K2-Think models before and after RL fine-tuning described in Section\ref{sec:experiments}:
K2-Think-SFT sees compilation error (L1+L2) drop 10.8\%, functional error (L3S+L3U) rise 3.5\%, and Pass rise 7.3\%.
K2-Think sees compilation error (L1+L2) drop 8.6\%, functional error (L3S+L3U) rise 2.6\%~pp, and Pass rise 6.0\%.
The increase of L3S in K2-Think-SFT and K2-Think after RL fine-tuning is more than the increase in L3U, indicating that more propagated errors from L1 and L2 are becoming solvable via inference-time scaling.
However, L3U remains the dominant L3 category post-RL, confirming that RL amplifies existing competencies without creating new hardware understanding.
Comparing CodeV-R1-Distill-7B (SFT only) to CodeV-R1-Qwen-7B (SFT+RL) from~\citet{zhu2026qimengcodevr1}, the additional RL phase yields a modest +3.4~pp pass-rate improvement while L3 remains above 27\%, reinforcing the pattern that RL after SFT yields diminishing returns for functional correctness.

\paragraph{(7)~Complementary knowledge: most L3U errors are model-specific.}
Although every individual model exhibits persistent L3U errors, the \emph{union} of all 17 evaluated models collectively solves 150 of the 156 VerilogEval-Human problems (96.2\%).
The best single model solves 148 problems, and individual L3U counts range from 8 (Gemini~3.1~Pro) to 111 (Qwen2.5-Coder-7B)---yet only 6 problems remain unsolvable by \emph{every} model.
This means the vast majority of any given model's L3U errors are \emph{not} intrinsically unsolvable; they are model-specific knowledge gaps that other models have filled through different pretraining data or training paradigms.
The finding implies that model diversity---through ensembles, routing, or complementary pretraining corpora---could close most of the L3U gap without requiring breakthroughs in model architecture.
The remaining 6 universally unsolvable problems (marked with $\dagger$ in Table~\ref{tab:always_failed} and Table~\ref{tab:always_failed_p2}) span Sequential\_Logic and Mux\_Select types, and involve complex multi-step temporal reasoning or non-standard specifications that no current approach masters.

Additional discussion about RTL problem-type analysis across all models can also be found in Section~\ref{sec:per_type_analysis}.
\section{When Language Models Fail to Learn RTL Coding}
\label{sec:experiments}

A second key question motivating this work is: \emph{can RL fine-tuning help language models generalize prior knowledge from other domains to the hardware domain?}
The K2-Think models~\cite{k2think2025} used in our experiments are pretrained and instruction-tuned using exclusively non-RTL data~\cite{k2thinksft2025}, making GRPO fine-tuning on RTL tasks a direct test of cross-domain generalization---whether RL can transfer coding competencies acquired from general-purpose programming to hardware design.

We apply the fault analysis framework defined in Section~\ref{sec:fault} to study how this RL fine-tuning affects error distributions.
We train two GRPO configurations on K2-Think models and track L1/L2/L3S/L3U/Pass rates across training and validation. A complementary problem-type discussion is provided in Section~\ref{sec:empirical} to further explore the training/validation performance in terms of different hardward design categories.

\paragraph{GRPO Training.}
GRPO~\cite{shao2024deepseekmath} samples $K$=32 rollouts per prompt, computes rewards, and updates the policy toward higher-reward responses relative to a group baseline.
The reward function combines a format component ($r_\text{fmt}$=0.1, verifying valid output structure) and a testbench component ($r_\text{tb}$=1.0, checking functional correctness via simulation).
Each training step uses 8~prompts sampled from the training dataset disjoint from the validation set, yielding 256 rollouts per step. The training dataset consists of problem specifications and Verilog code snippets used by ScaleRTL~\cite{deng2025scalertl}.

\paragraph{Two Experiments.}
E1 (K2-Think-SFT) applies GRPO directly after general supervised fine-tuning.
E2 (K2-Think) additionally has general-domain RL pretraining before RTL-specific GRPO.
Both use the same reward scheme (0.1f + 1.0t) and produce explicit chain-of-thought via \texttt{<think>} blocks.
Table~\ref{tab:setup} summarizes detailed training configurations. Complementary discussion about the training data distribution can be found in Section~\ref{sec:type_exposure}

\paragraph{Validation.}
We evaluate on 156 VerilogEval-Human v1 problems~\cite{liu2023verilogeval} with 10 rollouts per problem (temperature 0.2), both before and after GRPO.
This yields 1{,}560 validation rollouts per experiment per phase.

\paragraph{Scale.}
We load all available training steps: 494 (E1) + 516 (E2) = 1{,}010 steps, totaling 258{,}560 rollouts.
Training dynamics are visualised via accumulated rates: at step $t$, each rate equals the cumulative count from step~0 to~$t$ divided by total rollouts over the same range.

\subsection{Error Distribution Shifts}
\label{sec:syntax_semantic}

\paragraph{Validation: Before vs.\ After RL.}
\begin{figure}[t]
\centering
\includegraphics[width=\columnwidth]{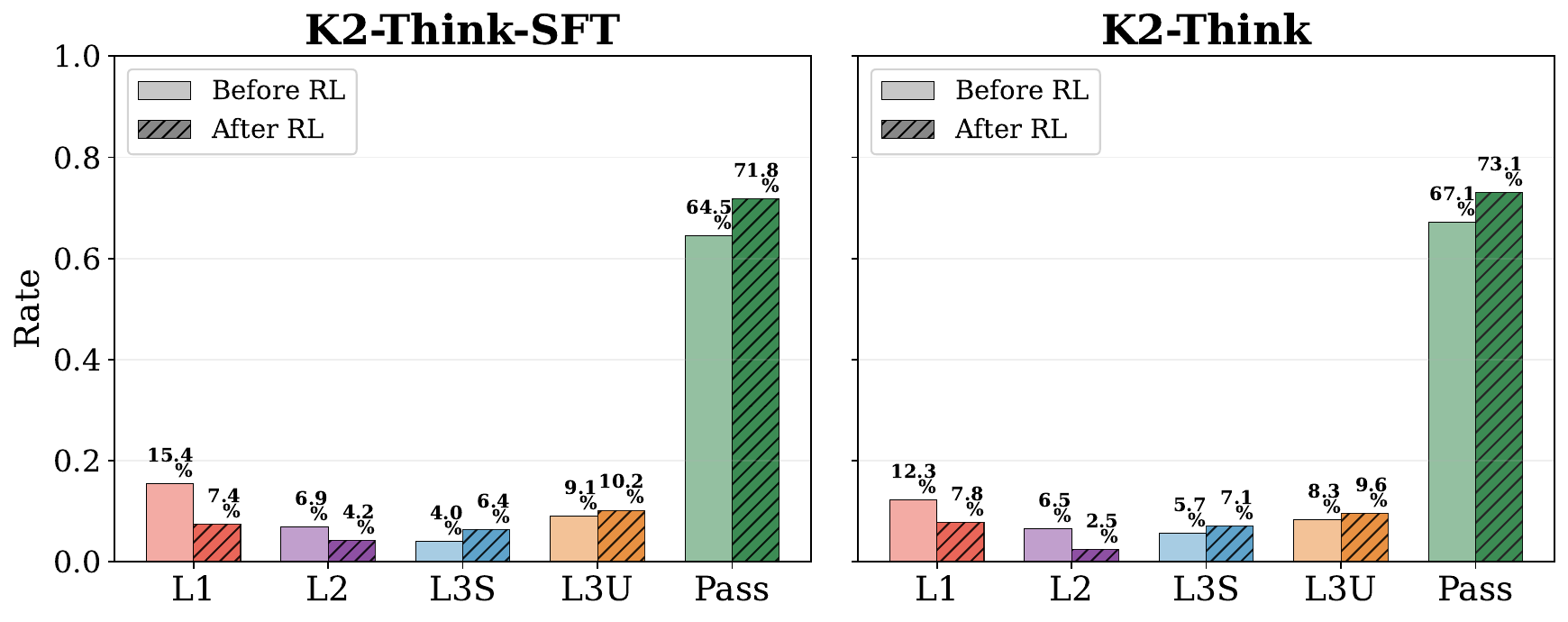}
\caption{Validation error rates before vs.\ after RL (grouped bars) on VerilogEval-Human V1 Benchmark. 
RL reduces L1(Syntactic) and L2(Semantic) errors but increases both L3S(Solvable) and L3U(Unsolvable) cases, indicating that newly compilable code shifts into functional failures.}
\label{fig:validation_bars}
\end{figure}

Figure~\ref{fig:validation_bars} shows validation error rates before and after GRPO for both experiments, decomposed into all four error levels.
The key pattern: \textbf{L1 (Syntactic) and L2 (Semantic) errors consistently drop, while both L3S (Solvable) and L3U (Unsolvable) failures consistently rise.}
For K2-Think-SFT (E1), L1 drops 8.0~pp (15.4\%$\to$7.4\%) and L2 drops 2.7~pp (6.9\%$\to$4.2\%) while the combined L3 rate rises 3.5~pp (13.1\%$\to$16.6\%), yielding a net pass-rate improvement of 7.3~pp (64.5\%$\to$71.8\%).
K2-Think (E2) shows a similar pattern: L1 drops 4.5~pp and L2 drops 4.0~pp while L3 rises 2.6~pp, with pass rate improving 6.0~pp (67.1\%$\to$73.1\%) at step~516.
Within L3, the increase is driven primarily by L3S~(Solvable): for K2-Think-SFT, L3S rises 2.4~pp (4.0\%$\to$6.4\%) while L3U rises 1.1~pp (9.1\%$\to$10.2\%); K2-Think shows a similar split (L3S +1.4~pp, L3U +1.3~pp).
This indicates that the newly compilable rollouts often fail on problems the model \emph{can} eventually solve, consistent with RL increasing the solvability rate (K2-Think-SFT: 78.2\%$\to$84.0\%) by converting some previously unsolvable problems into solvable ones (L3U$\to$L3S or Pass).

We hypothesise that the L1+L2 reduction \emph{causes} the L3 increase via selection pressure.
Before RL, many rollouts fail at L1 or L2 and never reach the testbench.
While these rollouts compile after RL, the model’s lack of full functional knowledge causes them to fail at L3 instead; these failures are distributed across L3S and L3U, with L3S absorbing the larger share.
The total number of ``wrong'' rollouts stays similar; they simply progress further through the error pipeline.

\begin{figure}[t]
\centering
\includegraphics[width=\columnwidth]{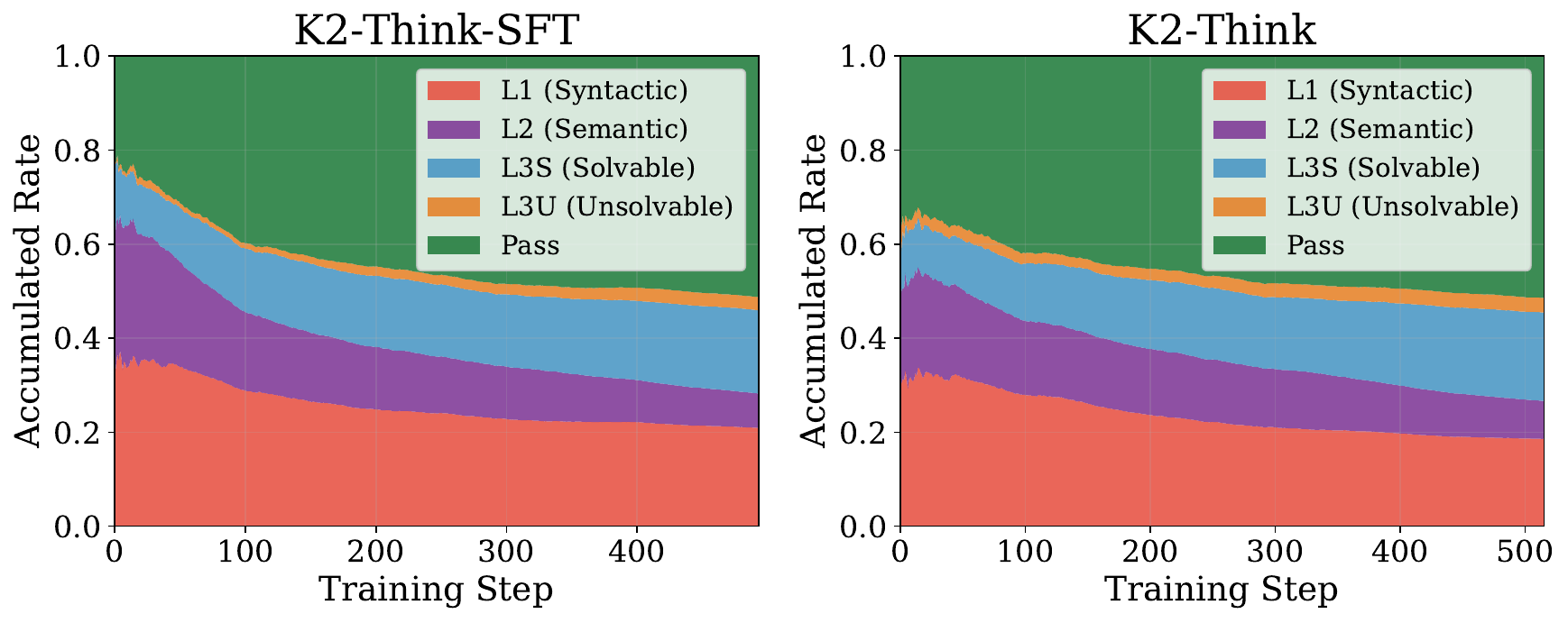}
\caption{Accumulated training error rates (stacked area). L1+L2+L3S+L3U+Pass=100\% at each step. Solvability is determined per-prompt from whether any rollout in the batch passes (Pass@K$>$0). L1 and L2 shrink rapidly; the L3 band---split into L3S and L3U---dominates the persistent error.}
\label{fig:accumulated_stacked}
\end{figure}

\paragraph{Training Dynamics.}

Figure~\ref{fig:accumulated_stacked} shows the accumulated error rates as stacked area charts over training, decomposed into L1, L2, L3S~(Solvable), L3U~(Unsolvable), and Pass.
Both experiments follow the same arc: accumulated L1+L2 drops steeply in early training---from $\sim$60\% to $\sim$46\% by step~100 for K2-Think-SFT~(E1)---and continues declining throughout the run.
K2-Think-SFT converges to $\sim$28\% accumulated L1+L2, $\sim$21\% L3 (dominated by L3S at $\sim$18\%, with L3U $\sim$3\%), and $\sim$51\% accumulated pass by step~493.
K2-Think~(E2) reaches similar final rates ($\sim$27\% L1+L2, $\sim$22\% L3, $\sim$51\% pass by step~515).
The L3 band grows slowly from $\sim$17\% to $\sim$21--22\% over training, absorbing some of the rollouts that previously failed at L1 or L2, consistent with the selection-pressure hypothesis above.

Because accumulated rates average over the full trajectory, including the poor early steps, the accumulated pass rate ($\sim$51\%) substantially underestimates the final model’s capability.
The final model's validation pass (71.8\% for E1, 73.1\% for E2) reflects the actual quality of the converged policy rather than the trajectory average.

\begin{figure}[t]
\centering
\includegraphics[width=\columnwidth]{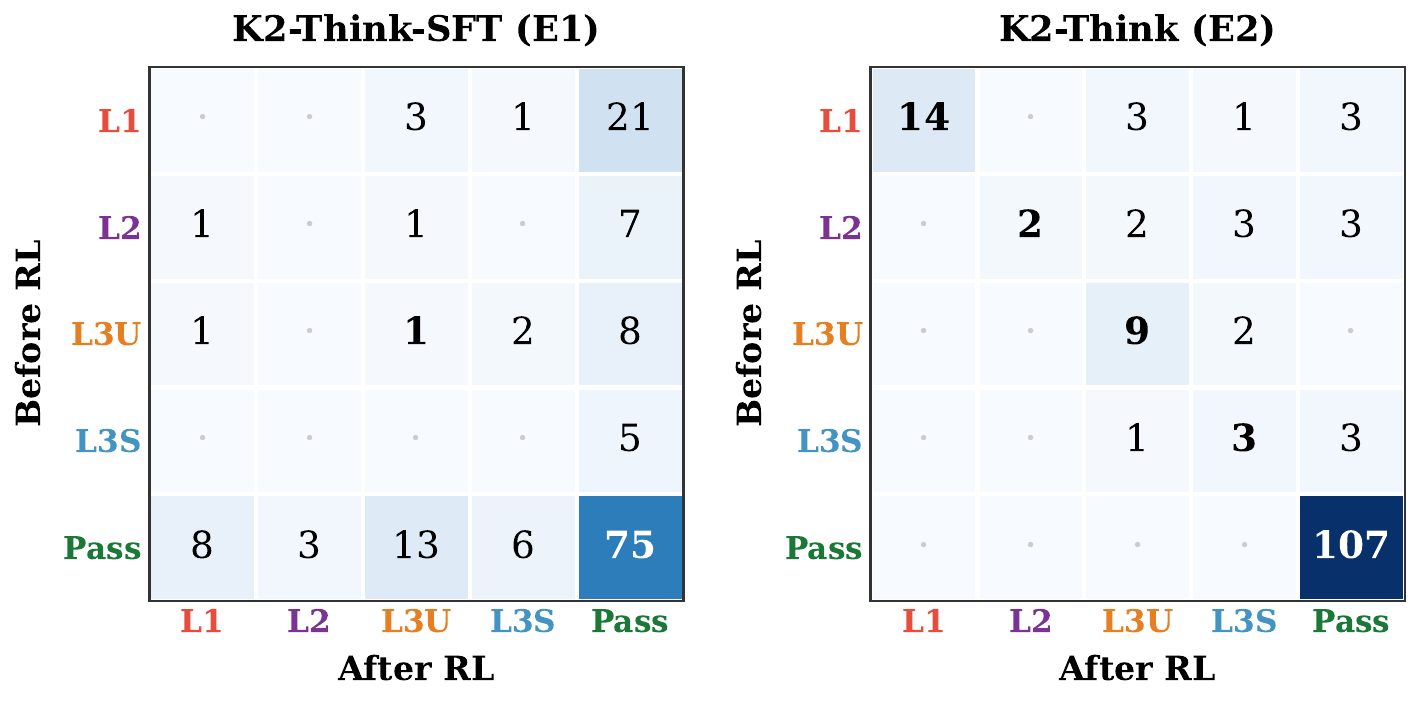}
\caption{Problem-level transition matrices (before vs.\ after RL). Each of the 156 VerilogEval problems is assigned a majority-vote category over its 10 rollouts. Cell~$(i,j)$ counts problems moving from error type~$i$ to~$j$. E2 (with general RL pretraining) shows near-perfect stability (107/107 Pass$\to$Pass), while E1 exhibits significant bidirectional movement including 13 Pass$\to$L3U regressions.}
\label{fig:transition_matrix}
\end{figure}

\paragraph{Problem-Level Transitions.}
To reveal how \emph{individual problems} migrate across error taxonomy defined in Section~\ref{sec:fault}, Figure~\ref{fig:transition_matrix} shows problem-level transition matrices for both experiments.
Each of the 156 validation problems is assigned a single category via majority vote over its 10 rollouts (ties broken by the lower pipeline level), and cell~$(i,j)$ counts problems whose majority category shifts from~$i$ before RL to~$j$ after RL.

The two experiments exhibit strikingly different transition profiles.
\textbf{K2-Think~(E2), which received general-domain RL pretraining, shows near-perfect stability}: all 107 majority-Pass problems remain Pass after RL, with only 2~regressions out of 156 (135~stable, 19~improved).
Improvements are distributed across categories, with L1$\to$Pass, L2$\to$Pass, and L3S$\to$Pass each contributing three problems, indicating broad capability gains.
In contrast, \textbf{K2-Think-SFT~(E1) exhibits substantial bidirectional movement}: 46~problems improve while 34~regress; notably, 13 shift from Pass to L3U, meaning problems the model reliably solved before RL become \emph{unsolvable} afterward.
The largest off-diagonal entry is L1$\to$Pass~(21), confirming that RL dramatically helps syntax-dominated problems.
However, the 30~Pass$\to$\{L1,L2,L3S,L3U\} regressions reveal that RL without prior general RL can also destabilise previously solved problems.

This contrast suggests that general-domain RL pretraining acts as a \emph{capability stabiliser}: E2 only adds new capabilities without losing existing ones, while E1 trades some existing capabilities for others.
This finding complements the aggregate rates in Figure~\ref{fig:validation_bars}. While both experiments yield similar net pass-rate gains (+7.3~pp vs. +6.0~pp), the underlying problem-level dynamics remain fundamentally different.

\section{Conclusion}
\label{sec:conclusion}

We evaluated nine proprietary frontier models, five open-source models, three RTL-specialized models, and two GRPO-finetuned K2-Think models on 156 VerilogEval problems in single-turn generation, applying a four-level error taxonomy (L1~syntactic, L2~semantic, L3S~solvable functional, L3U~unsolvable functional).
These results suggest that current LLMs' RTL coding ceiling is set primarily by pretraining knowledge.
Both SFT and RL move errors forward through the EDA pipeline rather than eliminating them: SFT teaches models to compile, revealing latent knowledge gaps at the functional testing stage, while RL amplifies existing competencies without creating fundamentally new hardware understanding.
Future directions include RTL-specific pretraining with hardware temporal reasoning patterns to breach the L3U ceiling and leveraging model diversity to exploit complementary knowledge across families.
Agentic approaches~\cite{zhang2025understanding,deng2026acertl} are also promising for attacking the ``universally hard'' RTL problems with iterative refinement.

\section*{Limitations}

Our analysis of language model failures and generalizations in RTL coding is subject to several methodological limitations.
First, the solvability classification (L3S vs.\ L3U) introduced in our fault analysis framework is inherently bounded by the number of rollouts $K$ (set to 10 for validation and 32 for training). A larger $K$ could shift some problems from Unsolvable (L3U) to Solvable (L3S), meaning our reported L3U rates represent an empirical upper bound on true unsolvability.
Second, all evaluations are strictly single-turn without iterative compilation feedback. While this isolates the models' internalized knowledge, it does not reflect multi-turn, agentic workflows that can automatically resolve many syntactic (L1) and semantic (L2) errors.
Third, our RL fine-tuning experiments are conducted on the open-source K2-Think model family to study cross-domain generalization via GRPO. The training dynamics and problem-level transition behaviors, including the trade-off between capability stabilization and regression, may differ for larger frontier models or those with RTL-specific pretraining.
Fourth, our L1/L2/L3 classification for training data rollouts relies on reward-signal heuristics and structural pattern matching rather than full EDA compilation, which may introduce noise into the accumulated error rate tracking.
Finally, while the VerilogEval-Human benchmark (156 problems) provides a rigorous test of fundamental RTL concepts, it primarily evaluates module-level design and may not fully represent the complexity of industrial-scale hardware systems.
.


%

\bibliography{custom}

\appendix

\section{Appendix}
\begin{table}[ht]
\centering
\caption{Two GRPO experiments on K2-Think models. Both use format + testbench reward (0.1f + 1.0t).}
\label{tab:setup}
\small
\begin{tabularx}{\columnwidth}{Xcc}
\toprule
& E1 & E2 \\
\midrule
Model & K2-Think-SFT & K2-Think \\
Reward & 0.1f+1.0t & 0.1f+1.0t \\
Steps & 494 & 516 \\
Prompts / step & 8 & 8 \\
Rollouts / prompt & 32 & 32 \\
Global batch size & 256 & 256 \\
Micro batch size & 1 & 1 \\
Max sequence length & 16384 & 16384 \\
Generation temp. & 1.0 & 1.0 \\
Learning rate & 1e-6 & 1e-6 \\
KL penalty & 0.001 & 0.001 \\
\bottomrule
\end{tabularx}
\end{table}
\section{Empirical Perspectives}
\label{sec:empirical}

\subsection{RTL Problem Type Classification}
\label{sec:problem_types}

We classify each VerilogEval problem into one of nine hardware design categories based on keyword matching on the problem specification:

\begin{enumerate}[leftmargin=*,itemsep=1pt]
    \item \textbf{FSM} (32): state machines, Mealy/Moore machines, state encoding
    \item \textbf{Counter\_Timer} (13): up/down counters, BCD, timers, thermostats
    \item \textbf{Sequential\_Logic} (23): flip-flops, latches, edge detectors, clock dividers, UART/SPI protocols
    \item \textbf{Shift\_Rotate} (11): LFSRs, barrel shifters, rotators, SIPO/PISO
    \item \textbf{Arithmetic} (8): adders, multipliers, ALU, CRC, parity, popcount
    \item \textbf{Encoder\_Decoder} (3): priority encoders, 7-segment decoders, scancode mapping
    \item \textbf{Mux\_Select} (10): multiplexers, demultiplexers, crossbar switches
    \item \textbf{Combinational\_Logic} (40): Karnaugh maps, truth tables, gates, waveform-based circuits, comparators
    \item \textbf{Wire\_Vector} (15): wire connections, vector manipulation, bit reversal, sign extension, constants
\end{enumerate}

Problems that do not match any category are labelled \textbf{Other} (1 of 156).
Among the 156 VerilogEval-Human problems, Combinational\_Logic (40) and FSM (32) are the most populated categories.

\subsection{Exposure--Performance Mismatch}
\label{sec:type_exposure}
\begin{figure}[t]
\centering
\includegraphics[width=\columnwidth]{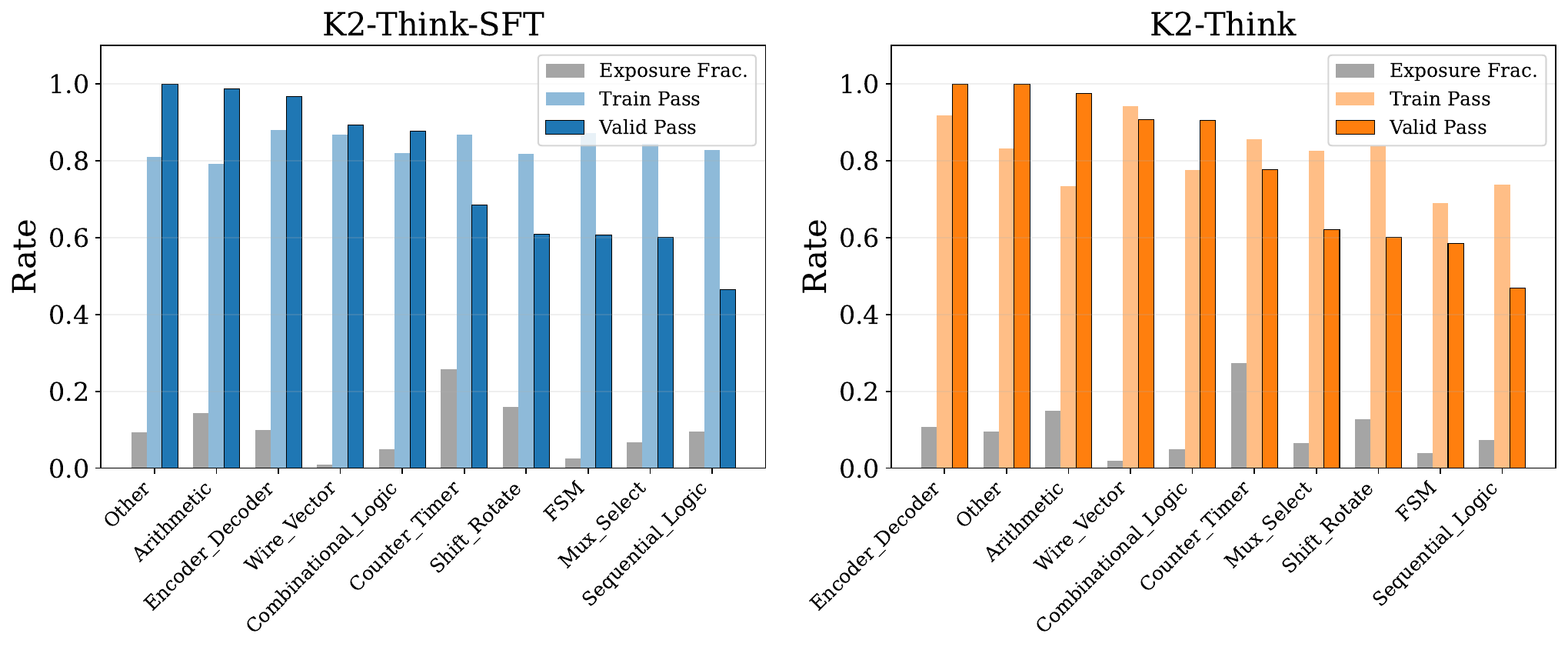}
\caption{Training exposure vs.\ performance by problem type defined in Section~\ref{sec:problem_types}. Per-experiment bar chart showing training exposure fraction (gray), training pass rate (light), and validation pass rate (dark). High exposure does not predict high validation performance.}
\label{fig:exposure_bars}
\end{figure}

Figure~\ref{fig:exposure_bars} shows, for each experiment, the training exposure fraction alongside training and validation pass rates per problem type.
No positive correlation between exposure and performance is visible:

\begin{itemize}[leftmargin=*,itemsep=1pt]
    \item \textbf{FSM}: high training exposure ($\sim$25\%) but low validation pass (58--61\%).
    \item \textbf{Combinational\_Logic}: moderate exposure ($\sim$15\%), high pass (88--91\%).
    \item \textbf{Arithmetic}: 14--16\% exposure, 97--99\% pass.
    \item \textbf{Wire\_Vector}: low exposure ($\sim$5\%), high pass (89--91\%).
\end{itemize}
Wire\_Vector receives far less training exposure than FSM yet achieves much higher pass rates.
This demonstrates that RL exposure cannot compensate for lack of pretraining knowledge---performance is driven primarily by the knowledge already encoded during pretraining.

\begin{table*}[t!]
\centering
\caption{Per-type pass rates (\%) on VerilogEval-Human (156 problems).
Problem counts per type: FSM (32), C/T (13), Seq (23), Sh/R (11), Arith (8), E/D (3), Mux (10), Comb (40), W/V (15).
Further details on problem types can be found in \S\ref{sec:problem_types}.}
\label{tab:per_type_pass}
\scriptsize
\resizebox{\textwidth}{!}{
\begin{tabular}{llccccccccc}
\toprule
Type & Model & FSM & C/T & Seq & Sh/R & Arith & E/D & Mux & Comb & W/V \\
\midrule
 \multirow{9}{*}{Proprietary} & Claude Opus 4.6 & 95.6 & 92.3 & 71.7 & 95.5 & 100.0 & 100.0 & 79.0 & 93.8 & 98.0 \\
  & GPT-5.3 Codex & 90.3 & 99.2 & 68.7 & 98.2 & 97.5 & 100.0 & 80.0 & 91.8 & 93.3 \\
  & Gemini 3.1 Pro & 90.9 & 96.2 & 69.6 & 89.1 & 98.8 & 100.0 & 78.0 & 81.5 & 99.3 \\
  & GPT-5.4 & 68.1 & 97.7 & 66.1 & 99.1 & 98.8 & 86.7 & 80.0 & 83.0 & 94.0 \\
  & GPT-5.2 & 58.8 & 88.5 & 65.7 & 80.0 & 92.5 & 100.0 & 72.0 & 82.8 & 93.3 \\
  & Claude Sonnet 4.6 & 77.2 & 90.0 & 57.0 & 82.7 & 87.5 & 53.3 & 52.0 & 86.8 & 71.3 \\
  & GPT-OSS-120B & 57.8 & 80.0 & 57.0 & 70.9 & 63.8 & 83.3 & 72.0 & 83.2 & 65.3 \\
  & GPT-5.1 & 50.6 & 80.0 & 54.8 & 72.7 & 93.8 & 66.7 & 77.0 & 69.2 & 85.3 \\
  & Gemini 3 Pro & 80.6 & 76.2 & 42.6 & 60.9 & 86.2 & 90.0 & 58.0 & 64.2 & 41.3 \\
\midrule
 \multirow{3}{*}{RTL-special.} & CodeV-R1-Distill-7B & 47.2 & 71.5 & 52.6 & 64.5 & 100.0 & 66.7 & 67.0 & 75.0 & 82.0 \\
  & CodeV-R1-Qwen-7B & 48.1 & 73.1 & 57.4 & 78.2 & 100.0 & 63.3 & 70.0 & 78.2 & 88.0 \\
  & ScaleRTL-Qwen-32B & 59.7 & 80.0 & 52.2 & 74.5 & 100.0 & 76.7 & 67.0 & 89.2 & 90.7 \\
\midrule
 \multirow{5}{*}{Open-source} & Qwen2.5-Coder-7B & 3.1 & 3.1 & 13.9 & 0.9 & 36.2 & 26.7 & 14.0 & 20.0 & 3.3 \\
  & Qwen2.5-Coder-32B & 7.2 & 20.8 & 18.7 & 10.9 & 36.2 & 50.0 & 7.0 & 13.8 & 13.3 \\
  & DS-R1-Distill-32B & 22.2 & 57.7 & 35.7 & 39.1 & 78.8 & 83.3 & 52.0 & 67.2 & 56.0 \\
  & K2-Think-SFT & 44.1 & 60.0 & 43.9 & 47.3 & 93.8 & 86.7 & 59.0 & 84.2 & 84.7 \\
  & K2-Think & 46.9 & 64.6 & 46.1 & 44.5 & 96.2 & 93.3 & 58.0 & 88.0 & 88.7 \\
\midrule
 \multirow{2}{*}{RL-finetuned} & K2-Think-SFT & 60.6 & 68.5 & 46.5 & 60.9 & 98.8 & 96.7 & 60.0 & 87.8 & 89.3 \\
  & K2-Think & 58.4 & 77.7 & 47.0 & 60.0 & 97.5 & 100.0 & 62.0 & 90.5 & 90.7 \\
\bottomrule
\end{tabular}
}
\end{table*}

\subsection{Per-Type Pass Rate Analysis}
\label{sec:per_type_analysis}

Table~\ref{tab:per_type_pass} presents per-type pass rates across all model families.
\textbf{(1)~Persistent difficulty on FSM and Sequential\_Logic.}
Even the best model (Claude Opus 4.6, 90.8\% overall) achieves only 71.7\% on Sequential\_Logic problems; weaker models drop below 50\%.
FSM pass rates span 50.6\%--95.6\% across frontier models, confirming these problem types as the primary bottleneck.
\textbf{(2)~Near-universal success on Arithmetic, Wire/Vector, and Encoder/Decoder.}
Most frontier models achieve 85--100\% pass on these categories, indicating that basic computation and data-path tasks are well within current LLM capabilities.
\textbf{(3)~Consistent per-type difficulty across families.}
Across all model families, the per-type difficulty ranking is remarkably stable: FSM and Sequential\_Logic are universally hard, while Arithmetic and Wire\_Vector are universally easy.
The gap between open-source and frontier models is largest on FSM and Sequential\_Logic, and smallest on Arithmetic and Wire\_Vector, suggesting that the difficulty is intrinsic to the problem types and that tasks sharing algorithmic structure with general programming are more accessible regardless of model scale or specialisation.

\subsection{Pass@k: No Exploration--Exploitation Tradeoff}

Table~\ref{tab:pass_at_k} shows that RL improves pass@1, pass@5, and pass@10 simultaneously in both experiments---there is no greedy-vs-diversity tradeoff.
K2-Think-SFT (E1) improves pass@1 from 0.645 to 0.718 ($+$0.073), and K2-Think (E2) from 0.671 to 0.731 ($+$0.060), with simultaneous gains at pass@5 and pass@10.

\begin{table}[t]
\centering
\caption{Pass@k Comparison Before and After RL}
\label{tab:pass_at_k}
\resizebox{\columnwidth}{!}{
\begin{tabular}{l ccc ccc}
\toprule
& \multicolumn{3}{c}{Before RL} & \multicolumn{3}{c}{After RL} \\
\cmidrule(lr){2-4} \cmidrule(lr){5-7}
Model & p@1 & p@5 & p@10 & p@1 & p@5 & p@10 \\
\midrule
K2-Think-SFT & 0.645 & 0.761 & 0.782 & 0.718 & 0.819 & 0.840 \\
K2-Think & 0.671 & 0.793 & 0.827 & 0.731 & 0.824 & 0.853 \\
\bottomrule
\end{tabular}
}
\end{table}

\subsection{RL Helps Hard Problems Most---But Not the Hardest}

\begin{figure}[t]
\centering
\includegraphics[width=\columnwidth]{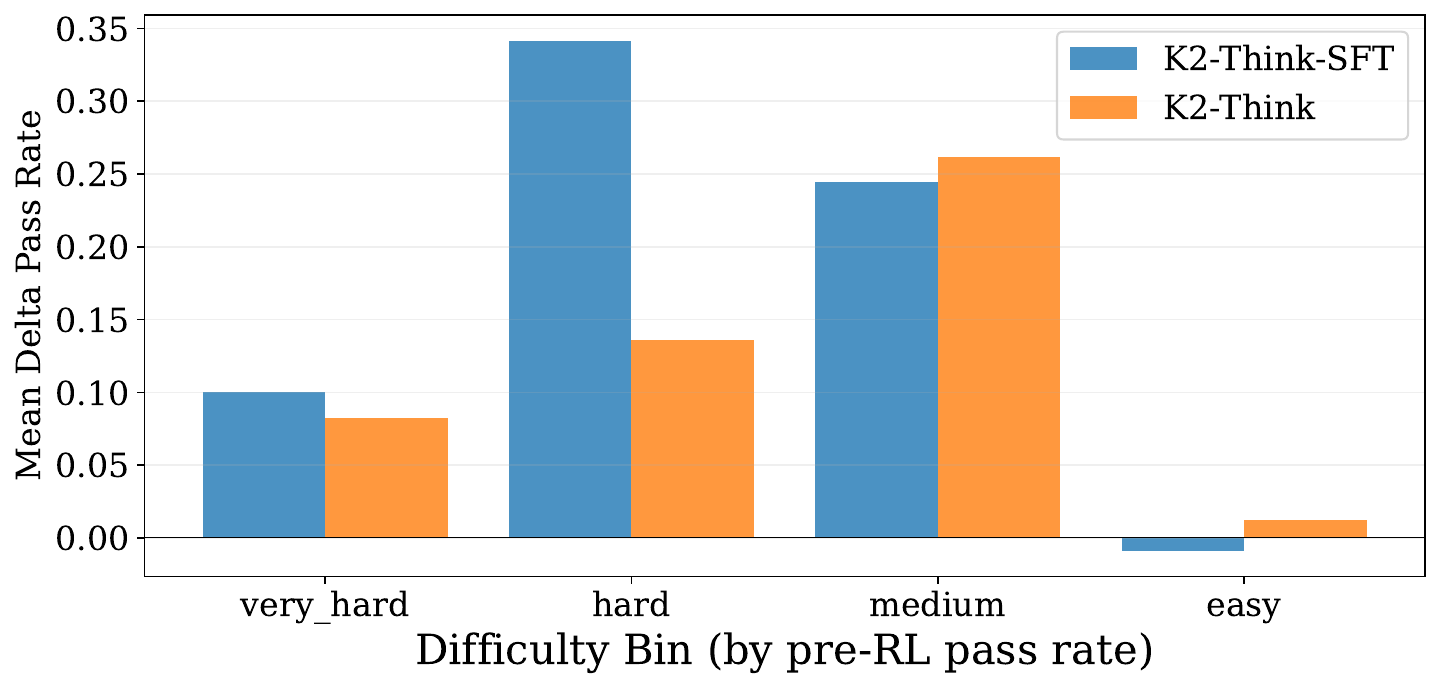}
\caption{RL improvement stratified by problem difficulty (validation). The largest gains are on initially-hard problems, but 10 problems never pass.}
\label{fig:difficulty_delta}
\end{figure}

We stratify the 156 validation problems by their pre-RL pass rate into four difficulty bins:
\textbf{very hard} (pass rate $\in [0, 0.2)$), \textbf{hard} ($[0.2, 0.5)$), \textbf{medium} ($[0.5, 0.8)$), and \textbf{easy} ($[0.8, 1.0]$).
Figure~\ref{fig:difficulty_delta} shows the mean change in pass rate (after RL minus before RL) computed on validation results for each bin.

Very hard problems see the largest mean improvement, while easy problems see minimal change.
The very hard bin is dominated by FSM and Sequential\_Logic types, whereas the easy bin contains mostly Arithmetic and Wire\_Vector.
However, a core set of problems remains at 0\% pass across all evaluation conditions (both experiments $\times$ both phases), indicating a hard ceiling that RL cannot breach.

\subsection{Universally Hard Problems}

Table~\ref{tab:always_failed} (Appendix) characterises the universally hard problems that remain at 0\% pass across both K2 experiments.
These span FSM, Mux\_Select, Shift\_Rotate, and Combinational\_Logic types.
Notable entries include \texttt{lfsr32} (a 32-bit linear feedback shift register), \texttt{rule110} (a cellular automaton classified as FSM), and \texttt{ece241\_2013\_q4} (a water reservoir FSM).
These problems require reasoning about multi-step temporal behaviour or non-standard specifications that no model masters.
Interestingly, frontier models also struggle with many of these problems (Table~\ref{tab:per_type_pass}), confirming that the difficulty is inherent to the problem rather than specific to the K2 model family.

\subsection{Exploration vs.\ Exploitation Dynamics}

We define \textbf{rollout diversity} to measure how structurally varied the generated solutions are within each prompt's rollout batch.
For each prompt at a given training step, we extract a \emph{structural fingerprint}: after stripping comments and whitespace, we extract the ordered sequence of control-flow keywords (\texttt{always}, \texttt{assign}, \texttt{if}, \texttt{case}, \texttt{for}, etc.) and join them as a string.
Rollout diversity is the ratio of unique fingerprints to total rollouts for a prompt, averaged across all prompts in a step.

\begin{figure}[t]
\centering
\includegraphics[width=\columnwidth]{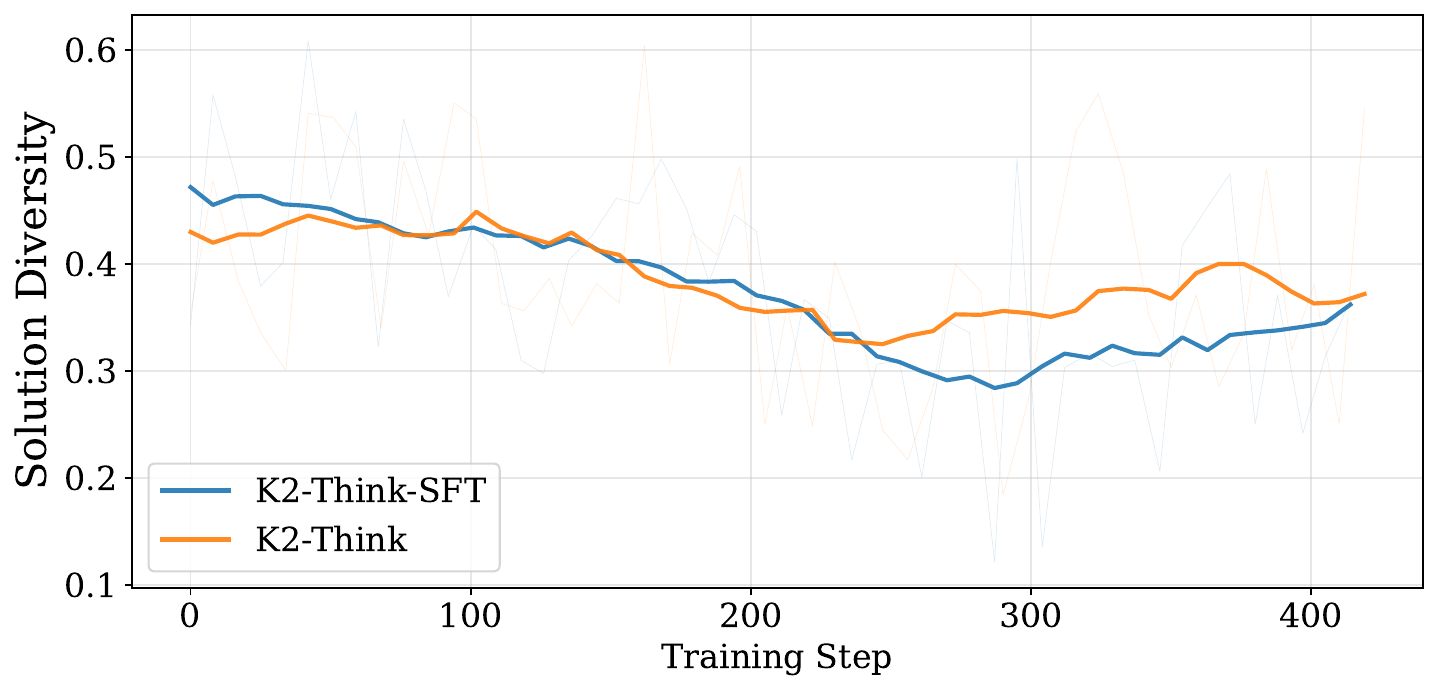}
\caption{Exploration vs.\ exploitation: solution diversity over training (moving-average smoothed). Diversity is defined as the ratio of unique structural fingerprints to total rollouts per prompt. Both K2 models maintain stable diversity ($\sim$0.34--0.37).}
\label{fig:diversity}
\end{figure}

Figure~\ref{fig:diversity} tracks this metric over training (smoothed with moving average, window 15).
Both K2 models maintain relatively stable diversity ($\sim$0.34--0.37) throughout training, with neither strong exploration nor exploitation trends.
This suggests that GRPO maintains a healthy balance between solution variety and convergence for thinking models.

\subsection{Training-Time Reward Consistency}

Reward variance analysis on the final training step reveals that both K2 experiments have predominantly mixed-outcome prompts.
Neither experiment has any all-fail prompts at its final step, confirming that RL successfully drives training-time reward to positive---the problem is not training failure but generalisation failure.
Both experiments have $<$1\% never-rewarded prompts, indicating that the format+testbench reward scheme provides sufficient learning signal.

\subsection{Solvability Analysis}
\label{sec:solvability}

The L3S/L3U split---whether at least one of the $K$=10 rollouts passes---provides a direct measure of a model's \emph{potential} on each problem, separating problems the model can solve stochastically from problems it fundamentally cannot.

\paragraph{Solvability rates across model families.}
Frontier models achieve the highest solvability rates: Claude~Opus~4.6 reaches 93.6\% (146/156 problems solvable), meaning that for only 10 problems does no rollout pass.
GPT-5.4 and GPT-5.1 reach 89.7\% and 89.1\% respectively.
Among RTL-specialized models, ScaleRTL-Qwen-32B achieves 89.1\%, comparable to GPT-5.1.
Open-source models span a wide range: K2-Think reaches 78.2\% pre-RL and 84.0\% post-RL, while Qwen2.5-Coder-7B is as low as 28.8\%.

\paragraph{Inference-time compute scaling.}
L3S errors are, by definition, recoverable via best-of-$N$ sampling: the model can produce a correct solution, so sampling more rollouts increases the probability of success.
L3U errors are \emph{not} recoverable by sampling---no number of additional rollouts will help.
This distinction has direct practical implications: for a deployed code-generation system, the optimal strategy depends on whether the model is in the solvable or unsolvable regime.
L3S problems benefit from inference-time compute scaling; L3U problems require model improvement.

\paragraph{RL effect on solvability.}
GRPO training increases the solvability rate for K2-Think-SFT from 78.2\% to 84.0\% (+5.8~pp), converting roughly 9 previously unsolvable problems into solvable ones.
This manifests as a shift from L3U to either L3S (the model now sometimes solves the problem) or Pass (the model now consistently solves it).
The remaining L3U problems---concentrated on FSM and Sequential\_Logic types---represent persistent knowledge gaps involving edge conditions in state transitions, reset behaviour, or boundary-case arithmetic.
These are the highest-value targets for model improvement: a small amount of targeted training data addressing these specific patterns could convert them to Pass.

\begin{table*}[ht]
\centering
\caption{Characterization of Always-Failed Validation Problems. These problems remain at 0\% pass across both K2 experiments. Problems marked with $\dagger$ are \emph{universally unsolvable}: no rollout from any of the 17 evaluated models (frontier, open-source, and RTL-specialized) passes.}
\label{tab:always_failed}
\scriptsize
\resizebox{\textwidth}{!}{
\begin{tabular}{l l l p{12cm}}
\toprule
Index & Task ID & Type & Prompt \\
\midrule
4 & 2013\_q2bfsm & FSM & Consider a finite state machine that is used to control some type of motor. The FSM has inputs x and y, which come from the motor, and produces outputs f and g, which control the motor. There is also a clock input called clk and a reset input (synchronous, active low) called resetn. The FSM has to work as follows. As long as the reset input is asserted, the FSM stays in a beginning state, called state A. When the reset signal is de-asserted, then after the next clock edge the FSM has to set the output f to 1 for one clock cycle. Then, the FSM has to monitor the x input. When x has produced the values 1, 0, 1 in three successive clock cycles, then g should be set to 1 on the following clock cycle. While maintaining g = 1 the FSM has to monitor the y input. If y has the value 1 within at most two clock cycles, then the FSM should maintain g = 1 permanently (that is, until reset). But if y does not become 1 within two clock cycles, then the FSM should set g = 0 permanently (until reset).  Here is the module template for you to complete:  ``` module top\_module ( 	input clk, 	input resetn, 	input x, 	input y, 	output f, 	output g );  // Your code here  endmodule ``` \\
7 & 2014\_q3fsm & FSM & Consider a finite state machine with inputs s and w. Assume that the FSM begins in a reset state called A, as depicted below. The FSM remains in state A as long as s = 0, and it moves to state B when s = 1. Once in state B the FSM examines the value of the input w in the next three clock cycles. If w = 1 in exactly two of these clock cycles, then the FSM has to set an output z to 1 in the following clock cycle. Otherwise z has to be 0. The FSM continues checking w for the next three clock cycles, and so on. Use as few states as possible. Note that the s input is used only in state A, so you need to consider just the w input. Assume reset is active high synchronous.  Here is the module template for you to complete:  ``` module top\_module ( 	input clk, 	input reset, 	input s, 	input w, 	output reg z );  // Your code here  endmodule ``` \\
22($\dagger$) & bugs\_mux2 & Mux\_Select & Find the bug and fix this 8-bit wide 2-to-1 mux.  // module top\_module ( //     input sel, //     input [7:0] a, //     input [7:0] b, //     output out  );  //     assign out = (\~sel \& a) | (sel \& b);  // endmodule  Here is the module template for you to complete:  ``` module top\_module ( 	input sel, 	input [7:0] a, 	input [7:0] b, 	output reg [7:0] out );  // Your code here  endmodule ``` \\
24 & circuit10 & Sequential\_Logic & This is a sequential circuit. The circuit consists of combinational logic and one bit of memory (i.e., one flip-flop). The output of the flip-flop has been made observable through the output state.  // Read the simulation waveforms to determine what the circuit does, then implement it.  // time            clk             a               b               state           q                // 0ns             0               1               x               x               x                // 5ns             1               1               x               x               x                // 10ns            0               0               0               x               x                // 15ns            1               0               0               0               0                // 20ns            0               0               0               0               0                // 25ns            1               0               0               0               0                // 30ns            0               0               0               0               0                // 35ns            1               0               0               0               0                // 40ns            0               0               0               0               0                // 45ns            1               0               1               0               1                // 50ns            0               0               1               0               1                // 55ns            1               1               0               0               1                // 60ns            0               1               0               0               1                // 65ns            1               1               1               0               0                // 70ns            0               1               1               0               0                // 75ns            1               0               0               1               1                // 80ns            0               0               0               1               1                // 85ns            1               1               1               0               0                // 90ns            0               1               1               0               0                // 95ns            1               1               1               1               1                // 100ns           0               1               1               1               1                // 105ns           1               1               1               1               1                // 110ns           0               1               1               1               1                // 115ns           1               1               0               1               0                // 120ns           0               1               0               1               0                // 125ns           1               0               1               1               0                // 130ns           0               0               1               1               0                // 135ns           1               0               0               1               1                // 140ns           0               0               0               1               1                // 145ns           1               0               0               0               0                // 150ns           0               0               0               0               0                // 155ns           1               0               0               0               0                // 160ns           0               0               0               0               0                // 165ns           1               0               0               0               0                // 170ns           0               0               0               0               0                // 175ns           1               0               0               0               0                // 180ns           0               0               0               0               0                // 185ns           1               0               0               0               0                // 190ns           0               0               0               0               0          Here is the module template for you to complete:  ``` module top\_module ( 	input clk, 	input a, 	input b, 	output q, 	output state );  // Your code here  endmodule ``` \\
44($\dagger$) & dff8 & Sequential\_Logic & Create 8 D flip-flops. All DFFs should be triggered by the positive edge of clk.  Here is the module template for you to complete:  ``` module top\_module( 	input clk, 	input [7:0] d, 	output reg [7:0] q);  // Your code here  endmodule ``` \\
50 & ece241\_2013\_q2 & Combinational\_Logic & A single-output digital system with four inputs (a,b,c,d) generates a logic-1 when 2, 7, or 15 appears on the inputs, and a logic-0 when 0, 1, 4, 5, 6, 9, 10, 13, or 14 appears. The input conditions for the numbers 3, 8, 11, and 12 never occur in this system. For example, 7 corresponds to a,b,c,d being set to 0,1,1,1, respectively. Determine the output out\_sop in minimum sum-of-products form, and the output out\_pos in minimum product-of-sums form.    Here is the module template for you to complete:  ``` module top\_module ( 	input a, 	input b, 	input c, 	input d, 	output out\_sop, 	output out\_pos );  // Your code here  endmodule ``` \\
51 & ece241\_2013\_q4 & FSM & A large reservior of water serves several users. In order to keep the level of water succificently high, three sensors are placed vertically at 5-inch intervals. When the water level is above the highest sensor s[3], the input flow rate should be zero. When the level is below the lowest sensor s[1], the flow rate should be at maximum (both Nominal flow valve and Supplemental flow valve opened). The flow rate when the level is between the upper and lower sensors is determined by two factors: the water level and the level previous to the last sensor change. Each water level has a nominal flow rate associated with it as show in the table below. If the sensor change indicates that the previous level was lower than the current level, the flow rate should be increased by opening the Supplemental flow valve (controlled by dfr).  // Water Level | Sensors Asserted | Nominal Flow Rate Inputs to be Asserted // Above s[3] | s[1], s[2], s[3] | None // Between s[3] and s[2] | s[1], s[2] | fr1 // Between s[2] and s[1]  | s[1] | fr1, fr2 // Below s[1] | None | fr1, fr2, fr3 // Also include an active-high synchronous reset that resets the state machine to a state equivalent to if the water level had been low for a long time (no sensors asserted, and all four outputs asserted).  Here is the module template for you to complete:  ``` module top\_module ( 	input clk, 	input reset, 	input [3:1] s, 	output reg fr3, 	output reg fr2, 	output reg fr1, 	output reg dfr );  // Your code here  endmodule ``` \\
55($\dagger$) & ece241\_2014\_q3 & Mux\_Select & For the following Karnaugh map, give the circuit implementation using one 4-to-1 multiplexer and as many 2-to-1 multiplexers as required, but using as few as possible. You are not allowed to use any other logic gate and you must use \_a\_ and \_b\_ as the multiplexer selector inputs, as shown on the 4-to-1 multiplexer below.  //       ab // cd   00 01 11 10 //  00 | 0 | 0 | 0 | 1 | //  01 | 1 | 0 | 0 | 0 | //  11 | 1 | 0 | 1 | 1 | //  10 | 1 | 0 | 0 | 1 |  // Consider a block diagram with inputs 'c' and 'd' going into a module called "top\_module". This "top\_module" has four outputs, mux\_in[3:0], that connect to a four input mux. The mux takes as input \{a,b\} and ab = 00 is connected to mux\_in[0], ab=01 is connected to mux\_in[1], and so in. You are implementing in Verilog just the portion labelled "top\_module", such that the entire circuit (including the 4-to-1 mux) implements the K-map.   Here is the module template for you to complete:  ``` module top\_module ( 	input c, 	input d, 	output [3:0] mux\_in );  // Your code here  endmodule ``` \\
56($\dagger$) & ece241\_2014\_q4 & Sequential\_Logic & Given the finite state machine circuit described below, assume that the D flip-flops are initially reset to zero before the machine begins.  // Build this circuit in Verilog.  // Input x goes to three different two-input gates: a XOR, an AND, and a OR gate. Each of the three gates is connected to the input of a D flip-flop and then the flip-flop outputs all go to a three-input XNOR, whose output is Z. The second input of the XOR is its corresponding flip-flop's output, the second input of the AND is its corresponding flip-flop's complemented output, and finally the second input of the OR is its corresponding flip-flop's complementary output.  Here is the module template for you to complete:  ``` module top\_module ( 	input clk, 	input x, 	output z );  // Your code here  endmodule ``` \\
61 & edgedetect2 & Mux\_Select & For each bit in an 8-bit vector, detect when the input signal changes from one clock cycle to the next (detect any edge). The output bit should be set the cycle after a 0 to 1 transition occurs.  Here is the module template for you to complete:  ``` module top\_module( 	input clk, 	input [7:0] in, 	output reg [7:0] anyedge);  // Your code here  endmodule ``` \\

\bottomrule
\end{tabular}
}
\end{table*}
\begin{table*}[ht]
\centering
\caption{Characterization of Always-Failed Validation Problems. These problems remain at 0\% pass across both K2 experiments. Problems marked with $\dagger$ are \emph{universally unsolvable}: no rollout from any of the 17 evaluated models (frontier, open-source, and RTL-specialized) passes.}
\label{tab:always_failed_p2}
\scriptsize
\resizebox{\textwidth}{!}{
\begin{tabular}{l l l p{12cm}}
\toprule
Index & Task ID & Type & Prompt \\
\midrule
82 & gshare & Combinational\_Logic & Build a gshare branch predictor with 7-bit pc and 7-bit global history, hashed (using xor) into a 7-bit index. This index accesses a 128-entry table of two-bit saturating counters. The branch predictor should contain a 7-bit global branch history register. The branch predictor has two sets of interfaces: One for doing predictions and one for doing training. The prediction interface is used in the processor's Fetch stage to ask the branch predictor for branch direction predictions for the instructions being fetched. Once these branches proceed down the pipeline and are executed, the true outcomes of the branches become known. The branch predictor is then trained using the actual branch direction outcomes.  // When a branch prediction is requested (predict\_valid = 1) for a given pc, the branch predictor produces the predicted branch direction and state of the branch history register used to make the prediction. The branch history register is then updated (at the next positive clock edge) for the predicted branch.  // When training for a branch is requested (train\_valid = 1), the branch predictor is told the pc and branch history register value for the branch that is being trained, as well as the actual branch outcome and whether the branch was a misprediction (needing a pipeline flush). Update the pattern history table (PHT) to train the branch predictor to predict this branch more accurately next time. In addition, if the branch being trained is mispredicted, also recover the branch history register to the state immediately after the mispredicting branch completes execution. // If training for a misprediction and a prediction (for a different, younger instruction) occurs in the same cycle, both operations will want to modify the branch history register. When this happens, training takes precedence, because the branch being predicted will be discarded anyway. If training and prediction of the same PHT entry happen at the same time, the prediction sees the PHT state before training because training only modifies the PHT at the next positive clock edge. The following timing diagram shows the timing when training and predicting PHT entry 0 at the same time. The training request at cycle 4 changes the PHT entry state in cycle 5, but the prediction request in cycle 4 outputs the PHT state at cycle 4, without considering the effect of the training request in cycle 4. Reset is asynchronous active-high.  Here is the module template for you to complete:  ``` module top\_module( 	input clk, 	input areset,   	input predict\_valid, 	input [6:0] predict\_pc, 	output predict\_taken,  	output [6:0] predict\_history,  	input train\_valid, 	input train\_taken, 	input train\_mispredicted, 	input [6:0] train\_history,  	input [6:0] train\_pc );  // Your code here  endmodule ``` \\
92 & lemmings4 & FSM & The game Lemmings involves critters with fairly simple brains. So simple that we are going to model it using a finite state machine. In the Lemmings' 2D world, Lemmings can be in one of two states: walking left (walk\_left is 1) or walking right (walk\_right is 1). It will switch directions if it hits an obstacle. In particular, if a Lemming is bumped on the left (by receiving a 1 on bump\_left), it will walk right. If it's bumped on the right (by receiving a 1 on bump\_right), it will walk left. If it's bumped on both sides at the same time, it will still switch directions.  // In addition to walking left and right and changing direction when bumped, when ground=0, the Lemming will fall and say ""aaah!"". When the ground reappears (ground=1), the Lemming will resume walking in the same direction as before the fall. Being bumped while falling does not affect the walking direction, and being bumped in the same cycle as ground disappears (but not yet falling), or when the ground reappears while still falling, also does not affect the walking direction. // In addition to walking and falling, Lemmings can sometimes be told to do useful things, like dig (it starts digging when dig=1). A Lemming can dig if it is currently walking on ground (ground=1 and not falling), and will continue digging until it reaches the other side (ground=0). At that point, since there is no ground, it will fall (aaah!), then continue walking in its original direction once it hits ground again. As with falling, being bumped while digging has no effect, and being told to dig when falling or when there is no ground is ignored. (In other words, a walking Lemming can fall, dig, or switch directions. If more than one of these conditions are satisfied, fall has higher precedence than dig, which has higher precedence than switching directions.) // Although Lemmings can walk, fall, and dig, Lemmings aren't invulnerable. If a Lemming falls for too long then hits the ground, it can splatter. In particular, if a Lemming falls for more than 20 clock cycles then hits the ground, it will splatter and cease walking, falling, or digging (all 4 outputs become 0), forever (Or until the FSM gets reset). There is no upper limit on how far a Lemming can fall before hitting the ground. Lemmings only splatter when hitting the ground; they do not splatter in mid-air. // Implement a Moore state machine that models this behaviour. areset is positive edge triggered asynchronous reseting the Lemming machine to walk left.  Here is the module template for you to complete:  ``` module top\_module ( 	input clk, 	input areset, 	input bump\_left, 	input bump\_right, 	input ground, 	input dig, 	output walk\_left, 	output walk\_right, 	output aaah, 	output digging );  // Your code here  endmodule ``` \\
94 & lfsr5 & Shift\_Rotate & A linear feedback shift register is a shift register usually with a few XOR gates to produce the next state of the shift register. A Galois LFSR is one particular arrangement where bit positions with a "tap" are XORed with the output bit to produce its next value, while bit positions without a tap shift. If the taps positions are carefully chosen, the LFSR can be made to be "maximum-length". A maximum-length LFSR of n bits cycles through 2**n-1 states before repeating (the all-zero state is never reached). Build a 5-bit maximal-length Galois LFSR with taps at bit positions 5 and 3. The active-high synchronous reset should reset the LFSR output to 1.  Here is the module template for you to complete:  ``` module top\_module( 	input clk, 	input reset, 	output reg [4:0] q);  // Your code here  endmodule ``` \\
99($\dagger$) & m2014\_q4d & Sequential\_Logic & Implement in Verilog the following circuit: A D flip-flop takes as input the output of a two-input XOR. The flip-flop is positive edge triggered by clk, but there is no reset. The XOR takes as input 'in' along with the output 'out' of the flip-flop.  Here is the module template for you to complete:  ``` module top\_module ( 	input clk, 	input in, 	output logic out );  // Your code here  endmodule ``` \\
111($\dagger$) & mt2015\_muxdff & Sequential\_Logic & Consider this Verilog module "full\_module":  // module full\_module ( //     input [2:0] r, //     input L, //     input clk, //     output reg [2:0] q );  // always @(posedge clk) begin //     if (L) begin //         q <= r; //     end else begin //         q <= \{q[1] \^ q[2], q[0], q[2]\}; //     end // end  // endmodule  // You want to create a hierarchical Verilog design where a flipflop and 2-1 multiplexer are in a submodule, and that submodule is instantiated three times in this code. Create the submodule called "top\_module".   Here is the module template for you to complete:  ``` module top\_module( 	input clk, 	input L, 	input q\_in, 	input r\_in, 	output reg Q);  // Your code here  endmodule ``` \\
\bottomrule
\end{tabular}
}
\end{table*}

\end{document}